\documentclass[11pt,a4paper]{article}
\usepackage[hyperref]{acl2021}
\usepackage{inputenc}
\usepackage{times}
\usepackage{latexsym}
\usepackage{enumitem}
\usepackage{todonotes}
\usepackage{tcolorbox}
\usepackage{color,soul}
\usepackage{hyperref}
\usepackage{flushend}
\usepackage[section]{placeins}
\usepackage{epstopdf}
\usepackage{multirow,tabularx}
\usepackage{pbox}
\aclfinalcopy
\RequirePackage{colortbl}
\definecolor{airforceblue}{rgb}{0.36, 0.54, 0.66}
\definecolor{amaranth}{rgb}{0.9, 0.17, 0.31}
\definecolor{applegreen}{rgb}{0.55, 0.71, 0.0}
\definecolor{alizarin}{rgb}{0.82, 0.1, 0.26}
\definecolor{azure}{rgb}{0.0, 0.5, 1.0}
\definecolor{cadmiumgreen}{rgb}{0.0, 0.42, 0.24}

\title{Challenges and Considerations with Code-Mixed NLP for Multilingual Societies}

\author{Vivek Srivastava \\
  TCS Research\\ Pune, Maharashtra, India \\
  \texttt{srivastava.vivek2@tcs.com} \\\And
  Mayank Singh \\
  IIT Gandhinagar\\ Gandhinagar, Gujarat, India \\
  \texttt{singh.mayank@iitgn.ac.in} \\}

\begin{document}
\maketitle
\begin{abstract}
Multilingualism refers to the high degree of proficiency in two or more languages in the written and oral communication modes. It often results in language mixing, a.k.a. \textit{code-mixing},  when a multilingual speaker switches between multiple languages in a single utterance of a text or speech. This paper discusses the current state of the NLP research, limitations, and foreseeable pitfalls in addressing five real-world applications for social good---crisis management, healthcare, political campaigning, fake news, and hate speech---for multilingual societies. We also propose futuristic datasets, models, and tools that can significantly advance the current research in multilingual NLP applications for societal good. As a representative example, we consider English-Hindi code-mixing but draw similar inferences for other language pairs.
\end{abstract}

\section{Introduction}
The term  ``\textit{multilingualism}''  refers both to an individual's ability to use several languages and the coexistence of different language communities in one geographical area. Contrary to the general belief, most of the world's population is bilingual or multilingual. Monolingualism is characteristic only of a minority of the world's population~\cite{valdes2007multilingualism}. Multilingualism sometimes uses elements of multiple languages when conversing with each other, resulting in a \textit{code-mixed} conversation.  Code-mixing (aka \textit{code-switching}), as formally defined by~\citet{bokamba1989there}, is the mixing of words, phrases, and sentences from two distinct grammatical (sub)systems within the same speech event. With the wide-reaching popularity of social media platforms, code-mixing has emerged as one of the significant linguistic phenomena among multilingual communities~\cite{das2015code}. Throughout this paper,  we only consider code-mixing as intra-sentential wherein language switching occurs inside the sentence.

In this paper, we present five real-world applications for social good---\textit{crisis management, healthcare, political campaigning, fake news}, and \textit{hate speech }---and discuss the current ability of the NLP tools in handling code-mixed data. We critically analyze the progress in NLP in processing code-mixed data and present immediate research directions and potential pitfalls. In contrast to existing survey literature on code-mixing, we primarily focus on the NLP applications for societal good and evaluate NLP venues' contribution towards such applications from a code-mixing perspective. 

\begin{figure}[!t]
\centering
\small{
\begin{tcolorbox}[colback=white]

\textsc{\textcolor{alizarin}{English}}: We have a  fully autonomous vehicle.\\
\textsc{\textcolor{cadmiumgreen}{Romanized Hindi}}: hamaare paas poori tarah se svaayatt vaahan hai\\
\textsc{\textcolor{red}{Hinglish}}: Hamare paas fully autonomous vaahan hai
\end{tcolorbox}}
\caption{A representative Hinglish sentence and the corresponding parallel Hindi-English sentences~\cite{srivastava-singh-2020-phinc}.}
\label{fig:hinglish}
\end{figure}

\begin{table*}[!tbh]
\centering
\small{
\begin{tabular}{|c|l|}
\hline
  \textbf{Application} &\textbf{Example} \\ \hline
   \multirow{2}{*}{\pbox{2.1cm}{\centering Crisis Management}}  & \pbox{12.2cm}{\textbf{Hinglish:} Delhi NCR may bhukamp? Assam Bengal ke baad ab Delhi may}\\
  & \pbox{12.2cm}{\textbf{English:} Earthquake in Delhi NCR? After Assam Bengal now in Delhi.} \\\hline
  
   \multirow{2}{*}{\pbox{2.1cm}{\centering Healthcare}} & \pbox{12.2cm}{\textbf{Hinglish:} Nahi bole toh bhi kam se kam 250-300 nuske bata chuke hai yeh log Corona khatam karne ke liye. Aur WHO ainwayi keh raha hai ke Corona ke liye vaccine nahi bani abtak }\\
  & \pbox{12.2cm}{\textbf{English:} These people have suggested atleast 250-300 prescriptions to cure CORONA. And WHO, without any reason, is saying we have no vaccine for CORONA.}\\\hline

   \multirow{2}{*}{\pbox{2.3cm}{\centering Political Campaigning}} & \pbox{11cm}{\textbf{Hinglish:} ``Beti Bachao'' is a myth. ``Beti BJP Se Bachao'' is reality!}\\
  & \pbox{11cm}{\textbf{English:} Save daughter is a myth. Save daughter from BJP is a reality.}\\\hline
  
   \multirow{2}{*}{\pbox{2.3cm}{\centering Fake News}} & \pbox{12cm}{\textbf{Hinglish:} WHO ne Bharat ko Karuna free declare Kar Diya hai. Humare PM narinder Modi is being awarded best PM award by UNESCO}\\
  & \pbox{12cm}{\textbf{English:} WHO has declared India as Corona free. Our Prime minister Narindra Modi is being awarded best PM award by UNESCO..}\\\hline
  
 \multirow{2}{*}{\pbox{2cm}{Hate Speech
 }} & \pbox{11cm}{\textbf{Hinglish:} Look ye politicians suvar jaise baithe rahte hain sirf money ke liye kaam karte hain. They don’t care about public.}\\ 
  & \pbox{11cm}{\textbf{English:} Look these politicians are sitting like pigs and they only work for money. They don’t care about public.}\\
  \hline
\end{tabular}}

\caption{Representative examples for each application  as described in Section~\ref{sec:app}.}
\label{code-mixed-data}
\end{table*}

As a representative example, in this paper, we primarily focus on \textit{Hinglish}, a portmanteau of Hindi and English. Highlish is highly prevalent in the Indian sub-continent. \citet{baldauf2004hindi} projected in 2004 that the world's Hinglish speakers ($>$350 million) might soon outnumber native English speakers. Hinglish presents significant linguistic challenges. In contrast to other code-mixed pairs such as Spanglish (a portmanteau of Spanish and English), which carries a relative similarity in Spanish and English syntactic structure, Hinglish is constructed over two syntactically divergent languages~\cite{berk1986linguistic}.  Linguistic theories such as equivalence constraint theory~\cite{poplack1980sometimes} and size-of-constituent constraint theory~\cite{gumperz1969cognitive} and the associated NLP algorithms and tools that were proposed for syntactically similar languages poorly perform for a code-mixed text generated from syntactically divergent languages. Figure \ref{fig:hinglish} shows the Hinglish text and the corresponding parallel Hindi-English sentences.

We present a general introduction to the five applications for social good in the next section. Section~\ref{sec:current_SOTA} describes the current state of the research in these application areas. We present limitations of existing NLP research for code-mixing in Section~\ref{sec:limit} and identify futuristic application-specific datasets, models and tools in Section~\ref{sec:future}. Towards the end, Section~\ref{sec:conclusion} concludes the discussion. 

\section{Applications for societal good}
\label{sec:app}
This paper focuses on five popular applications for social good and discusses these applications from a multilingual community perspective, wherein the data generated comprises code-mixed text, largely available over the internet. Next, we describe each application in detail.

\subsection{Crisis Management} 
\label{sec:CM}
Crisis encompasses both man-made and natural events. Man-made crisis events include military war, explosions, acts of terrorism,  extreme pollution levels, hazardous material spills, fires, transportation accidents, structure failures, and mining accidents. In contrast, natural crisis events include catastrophic events with atmospheric, geological, and hydrological origins (e.g., droughts, earthquakes, floods, hurricanes, landslides)~\cite{PRASAD2017215}. Interestingly, the current COVID-19 crisis can be attributed to both man-made and natural crises.  

As expected, these crisis events affect a large population, resulting in a large volume of multi-modal data such as text messages, videos, and images. The generated text data includes real-time casualty and loss statistics, donation announcements, volunteering calls, relief camp locations, along with negative aspects such as misinformation, disinformation, and rumors during a health emergency.  Thus, the primary role of NLP systems in crisis data management is not only to enable a large proportion of the affected population to consume the factually correct data but also to reduce the effect of incorrect/false information before it reaches the affected population (hereafter, \textbf{crisis data misinformation (CDM)}). Another important NLP goal is disaster response, wherein the system identifies information related to emotional support to the affected population, donations and volunteering and caution and advice from government bodies and social organizations, government announcements related to infrastructure (hereafter, \textbf{crisis data response (CDR)}). The third major goal is to summarize the large temporal data volume in real-time and present it in a meaningful fashion for easier consumption (hereafter, \textbf{crisis data summarization (CDS)}). 

\subsection{Healthcare}
\label{sec:HC_desc}
The Healthcare domain has witnessed tremendous advancements in the last two decades. Many healthcare systems allow patients to access their electronic health record (EHR) notes online through patient portals.  Applications such as personalized healthcare chatbots are increasingly becoming popular~\cite{healthcare}. Online medical portals such as Prato\footnote{\url{http://www.practo.com}} and Plushcare\footnote{\url{https://plushcare.com}} are helping patients to connect and find doctors. Online healthcare blogs present well-rounded, perceptive, and informative news sources and opinions from healthcare experts. 

Understanding healthcare information from the above sources requires NLP at its core. The major goal of NLP is to understand the medical jargon and construct a mapping between non-standard terminologies used in different geographies. This understanding helps in constructing association between medical processes  (hereafter, \textbf{healthcare knowledge graph construction (HKG)}). The second major goal is in developing personalized healthcare chat agents (hereafter, \textbf{healthcare chatbots (HCB)}). The third major goal is to debunk the misinformation associated with the health and medical terms (hereafter, \textbf{healthcare misinformation (HMI)}). 

\subsection{Political campaigning}
Social platforms play a critical role in the new political campaigning paradigm. Politicians exchange views on the latest partisan developments and invite the public and citizens to comment, share ideas, and adhere to their political program~\cite{phillips2009online}. Previous research shows that the successful social media campaign positively correlates with the election outcomes~\cite{baxter2012does,barclay2015india,kanungo2015india,heredia2018impact}. Similarly, in the recent U.S. elections, a study~\cite{wharton} shows that within the first month of using Twitter, politicians were able to raise between 1\% and 3\% of what they would have raised in a traditional two-year campaign. Some of the positive aspects of social media-based political campaigns are wider reach, no traditional media filter, two-way discussions, and negligible funding requirements. Some of the negative aspects include malicious propaganda against rival parties and exaggerated achievements. 

NLP systems can be deployed to understand the ongoing political campaigns and to make people aware of the factually correct information and red-flag the misinformation (hereafter, \textbf{political data misinformation (PDM)}). Another important goal is to measure the public perception about the ongoing political campaigns and present it to the political parties and the general mass. These opinions help a political party to plan its programs and visions and a voter to see both sides of campaigns (hereafter, \textbf{political opinion extraction (POE)}). 

\subsection{Fake news}
Fake news is a critical yet challenging problem in NLP. The rapid rise of social networking platforms has yielded a vast increase in information accessibility and accelerated the spread of fake news~\cite{oshikawa2020survey}. We witness fake news spreaders on diverse platforms ranging from political discussions, movie recommendations, e-retail product reviews. In the last decade, we witness a surge in fake news on sensitive and critical topics like pandemic~\cite{kar2020no}, disasters~\cite{kwanda2020fake}, and war~\cite{salem2019fa}. In the ongoing pandemic scenario, where major business, education, and political activities have been confined to online settings, the volume of fake news has grown substantially~\cite{fakeonline2}.  

As the menace of fake news encompasses multiple domains, the primary goal of the NLP system is to detect such fake messages, posts, or reviews (hereafter, \textbf{fake news detection (FND)}). FND is highly challenging in a real-time setting, where incoming messages are flagged according to the likelihood of being fake. The second major goal is the course correction, i.e., to summarize and present facts against the fake news (hereafter, \textbf{fake news explanation (FNE)}).

\subsection{Hate speech}
Hate speech is commonly defined as any communication that disparages a person or a group based on some characteristic such as race, color, ethnicity, gender,  sexual orientation,  nationality, or religion~\cite{nockleby2000hate}. Due to the massive rise of user-generated web content, the amount of hate speech is also steadily increasing~\cite{schmidt2017survey}. Government organisations~\cite{hatespeech1,hatespeech2} have framed strict laws to deal with offensive messages sent through social media and online messaging applications. The major challenge in hate speech detection lies in the diversity of characteristics. An NLP system, depending on the characteristics of hate speech, should be able to process and detect such hateful content effectively (hereafter, \textbf{hate speech detection (HSD)}).  

\noindent To summarize, in this section, we elaborate on five applications of NLP for societal good. For each application, we propose multiple sub-tasks. Table~\ref{code-mixed-data} presents representative Hinglish examples for each application described in this section. Since we focus on the applications from a code-mixed data perspective, the next section describes the current NLP research status addressing the above issues in code-mixed data.

\section{The current state of the research}
\label{sec:current_SOTA}
We study two different perspectives to understand the current state of the computational linguistic research for the societal good applications in multilingual societies. First, we analyze the five major ACL and non-ACL conferences to understand the NLP community's global trend. Next, we present an analysis of the currently available resources and some top-performing systems for each of the five societal good applications.

\subsection{Are we organizing enough workshops for the societal good?}
Workshops are major sources to promote the research in a dedicated area of study. In this paper, we analyze five major NLP conferences (ACL, EMNLP, NAACL\footnote{NAACL happened in the year 2016, 2018, and 2019.}, COLING, and LREC\footnote{LREC happened in the year 2016, 2018, and 2020.}) organized in the past five years (2016--2020). Over the years, almost all the top preferred conferences in the NLP community are organizing workshops to promote and include community members' diverse research interests (see Table~\ref{tab:WS_count}).
Specifically, in the healthcare domain, we see a significant number of workshops in ACL, EMNLP, and LREC. For other societal good applications, the number of organized workshops is considerably low. Also, except the workshop on computational approaches to linguistic code-switching (CALCS),  none of these workshops primarily focus on code-mixed text.
\begin{table}[!tbh]
\resizebox{\hsize}{!}{
\begin{tabular}{|c|c|c|c|c|c|c|}
\hline
& \textbf{\parbox{2cm}{\centering Crisis\\ Management}} & \textbf{\parbox{2cm}{Healthcare}} & \textbf{\parbox{2.1cm}{\centering Political\\ Campaigning}} & \textbf{\parbox{1cm}{ \centering Fake News}} & \textbf{\parbox{1cm}{\centering  Hate Speech }} & \textbf{Total} \\ \hline
\textbf{ACL}&  3  &10&1 & 1&2  &82  \\ \hline
\textbf{EMNLP}  & 1 &8  &2 &3  & 3 & 65\\ \hline
\textbf{NAACL}  &  0  &  2  & 1&0  &0 & 51\\\hline
\textbf{COLING} &0 & 3& 1& 3& 2 & 40 \\ \hline
\textbf{LREC}& 0&7&3 & 2&2  & 96\\ \hline
\end{tabular}}
\caption{Distribution of the societal good workshops organized during the five-year interval (2016--2020) in the five major NLP conferences. Total represent total number of workshops in the five-year interval.}
\label{tab:WS_count}
\end{table}

\subsection{What is accomplished so far? Is it really ``state-of-the-art''?}
\label{sec:SOTA}
Here, we explore some major works in the five societal good applications for multilingual societies. We provide an empirical study to highlight various strengths and limitations of these notable works. Since we consider the majority of the popular and recent works in this discussion, it will help capture the community's diverse research interests. It will also pave the way for developing effective systems in the future.  

\subsubsection{Crisis management} 
In Section~\ref{sec:CM}, we discuss several NLP tasks in crisis and disaster management.  To the best of our knowledge, we do not find any work primarily focusing on code-mixed crisis data. However, we find several works that process noisy social-media text for generating insights and responses. We list these works below:
\begin{itemize}[noitemsep,nolistsep,leftmargin=*]
  \item \textbf{CDM}: We witness a large volume of work on addressing the crisis data misinformation. Particularly, majority of the recent works focus on curating misinformed social-media posts during COVID-19  pandemic~\cite{shahi2020fakecovid,li2020mm,kar2020no,patwa2020fighting,qazi2020geocov19}. Although all these works report multilingual data curation, we do not find any proposed misinformation identification model that leverages these datasets to train misinformation detection models. Additionally, we find several other monolingual datasets discussing events like Ukrainian conflict~\cite{doi:10.1080/17512786.2016.1163237}, Syrian War~\cite{salem2019fa} and Palu earthquake~\cite{kwanda2020fake}. The majority of the crisis datasets are monolingual and present limited opportunities to develop tools to identify multilingual misinformed content. This limitation fosters the code-mixed misinformation spreader to spread the false information in a time of crisis easily. 
  \item \textbf{CDR}: We witness very few works addressing crisis data response. For example, \citet{imran2013extracting} extracted valuable information nuggets relevant to disaster response from microblogs. They curated a small dataset of 1,233 English sentences. However, we do not find any work that curates the above extremely time-sensitive information available in the code-mixed language. 
  \item \textbf{CDS}: We witness few recent works on summarizing crisis-related data available as microblogs~\cite{rudra2015extracting,rudra2016summarizing,rudra2018identifying}. All these works address the problem in two phases. In the first phase, they extract the relevant situational information from among the large volume of tweets. In the second phase, they summarize the extracted information. Note that the proposed methods only work for the English Language. Off-the-shelf usage of these works for the code-mixed languages might not give the desired results.
\end{itemize}

\subsubsection{Healthcare}
In Section \ref{sec:HC_desc}, we discussed the importance and the necessities to address the various healthcare-related challenges. We layout three major sub-fields (HKG, HCB, and HMI) where computational linguistic could play a major role. We could not find any dedicated study pertaining to the code-mixed languages to the best of our knowledge. Here, we present some well-known works in other languages and draw insights from them that present future research directions in the code-mixing domain.  
\begin{itemize}[noitemsep,nolistsep,leftmargin=*]
  \item \textbf{HKG}: Healthcare knowledge graphs~\cite{lindberg1993unified,oliveira2017recommendation} are extremely useful in converting enormous healthcare data in the machine-understandable format. HKG is also useful in informing healthcare practitioners of the latest research results~\cite{sadeghi2019linking}. It has proven useful in multiple instances such as diagnosis prediction~\cite{choi2017gram,ma2018kame} and analytics~\cite{aasman2018knowledge}. Code-mixed data available on various social media platforms (e.g., Twitter and Facebook) and healthcare discussion forums (e.g., India medical hub\footnote{http://www.indiamedicalhub.com/} and patient\footnote{https://patient.info}) can be used to expand existing monolingual HKGs. 
  
  \item \textbf{HCB}: Healthcare chatbots are the future for query resolution~\cite{amato2017chatbots,bates2019health}, consultation~\cite{kowatsch2017text}, medical history analysis~\cite{divya2018self}, and treatment~\cite{cameron2018best,bibault2019healthcare,chaix2019chatbots}. The majority of the HCBs are designed for formal communication, making it difficult for the multilingual speakers to have a human-like conversation with these chatbots as they lack the human emotions~\cite{palanica2019physicians}.
  
  \item \textbf{HMI}: Misinformation in the healthcare domain can result in a serious negative impact on the well-being of the people~\cite{tasnim2020impact}. Recently, a wave of misinformation related to the COVID-19 pandemic has propagated and studied~\cite{brennen2020types,pennycook2020fighting,kouzy2020coronavirus}. Several domain-specific misinformation detection systems~\cite{cui2020deterrent,hou2019towards,bal2020analysing,cui2020coaid} have been proposed to check the healthcare-related misinformation. The majority of these works focus on the monolingual data available on various social media platforms. However, due to the popularity of code-mixing conversations in multilingual communities, the risk of misinformation spread increases manifold. Most healthcare-related information is generated in monolingual language at the source (hospitals, healthcare centers, and diagnostic labs). 
\end{itemize}

\subsubsection{Political campaigning} 
The recent thrust on social media and online news platforms has contributed significantly to the change of medium for political campaigning. These platforms greatly facilitate reaching the masses quickly, but it adds several challenges such as misinformation and hate speech. Here, we explore two sub-fields (PDM and POE) where computational linguistic could help improve the situation. Since we cannot find any study for the code-mixed languages, we will explore monolingual and multilingual works.  

\begin{itemize}[noitemsep,nolistsep,leftmargin=*]
  \item \textbf{PDM}: Misinformation during election campaigning misleads people with manipulated facts and rumors~\cite{badawy2018analyzing,cantarella2020does}.  Multiple works identify the spread of political misinformation on various platform such as WhatsApp~\cite{garimella2020images,machado2019study,resende2019analyzing}, Facebook~\cite{guess2019less,mena2020cleaning}, and Twitter~\cite{shin2017political,grinberg2019fake, neudert2017junk}. In order to address political misinformation, various fact-checking platforms have been introduced~\cite{hassan2015quest,myslinski2015method,rashkin2017truth,hassan2017claimbuster,thorne2018automated}. Almost all of the above works curate and classify misinformation in English language.
  
  \item \textbf{POE}: Extracting opinion from the political campaigning data could be extremely helpful in building efficient election result forecasting systems and real-time visualizations~\cite{kaschesky2011opinion}. Various works have explored multiple sources to extract people's political opinions such as Twitter~\cite{asghar2014political,joshi2016political,gull2016pre}, news sources (such as NPR, Mother Jones, and Politico)~\cite{bamman2015open} and YouTube \cite{kandias2013youtube,acharoui2020identifying}. 
\end{itemize}

\subsubsection{Fake news}  
Fake news is one of the biggest challenges on social media platforms. Over the years, it has been an active area of research for the computational linguistic community. Unfortunately, we could not find any work for the fake news propagation in the code-mixed languages. Here, we explore some prominent works in other languages for the two major subareas (FND and FNE). 
\begin{itemize}[noitemsep,nolistsep,leftmargin=*]
  \item \textbf{FND}: Detecting fake news and the fake news spreaders is the primary step in checking the propagation of fake news~\cite{shu2017fake,meel2019fake}. Over the years, various benchmark datasets have been proposed~\cite{wang2017liar,golbeck2018fake,nakamura2020fakeddit} to detect fake news on various platforms. Recently, FND for low-resource~\cite{hossain2020banfakenews,amjad2020data} and multilingual~\cite{schwarz2020emet,shahi2020fakecovid} languages have gained interest from the computational linguistic community.  
  \item \textbf{FNE}: Black-box models for the fake news detection adds to the complexity of identifying and filtering the fake news~\cite{reis2019explainable}. The reasoning, interpretability, and the explanation \cite{shu2019defend} of the fake news detection models are extremely important for websites with extensive outreach (e.g., social media and news websites). Real-time prediction and the visualization of the fake news and its interpretation~\cite{yang2019xfake} could be extremely useful in several cases for multilingual societies. 
\end{itemize}

\subsubsection{Hate speech} In contrast to the above societal good applications, hate speech detection in the code-mixed languages has been an active area of exploration~\cite{santosh2019hate,chopra2020hindi,kamble2018hate}. Here, we discuss two very popular datasets (for the code-mixed languages) by \newcite{bohra2018dataset} (hereafter, HD1) and \newcite{mandl2019overview}) (hereafter, HD2) to detect hate speech on social media platforms. HD1 presents a dataset of 4,575 Hinglish tweets annotated with either of the binary labels: \textit{hate speech} or \textit{normal speech}. HD2 is a fine-grained hate-speech detection dataset in Hinglish. Labels in HD2 can be used to identify \textit{hate speech}, \textit{offensive language}, and the \textit{profanity} in the 5,983 tweets. Even though these works address the data scarcity issue for the code-mixed languages, the quality and the quantity of the examples in the dataset remains debatable. The size of HD1 and HD2 datasets is significantly small as compared to HD datasets in other monolingual languages~\cite{davidson2017automated,mathew2020hatexplain}. 

\section{Limitations}
\label{sec:limit}
As discussed in Section \ref{sec:SOTA}, the majority of the research for good societal applications for multilingual societies has enormous opportunities. This section discusses some major limitations that text NLP researchers and practitioners encounter very often in code-mixing. \newcite{srivastava-singh-2020-phinc} highlights six major limitations with the machine translation systems on the code-mixed text. The majority of these limitations stand true in dealing with societal good applications as well. Here, we present a comprehensive list of challenges and limitations that need to be carefully addressed to build robust and efficient systems and an encouraging environment in the multilingual research community.

\begin{itemize}[noitemsep,nolistsep,leftmargin=*]
\item \textbf{Limited text processing tools}: The limited availability of the text processing tools build especially for the code-mixed language makes it extremely challenging to create large scale robust and efficient systems. Efficient tools with functionalities like POS tagging, named entity recognition, language identification, transliteration, key phrase extraction, and topic modeling are yet to be developed at a large scale. The scarcity of high-quality datasets for these tasks makes it a vicious cycle to build multilingual societies' solutions. 

\item \textbf{Identifying and filtering code-mixed data}: One of the major challenges with the code-mixed data is the coexistence with other languages. We rarely find any platform where people communicate solely in code-mixed languages. The non-availability of automatic filtering tools makes it extremely challenging to create a large-scale code-mixed corpus. In most cases, people employ humans in the loop to identify and filter the code-mixed data~\cite{kumar2018aggression,vyas2014pos,chandu2019code}, but this approach is highly time and cost extensive. 

\item \textbf{Human bias}: Inherent human bias is one of the most significant and dangerous challenges for most NLP systems. Code-mixed languages also suffer from such biases very often. In addition to the bias present in the data generated on the various platforms (Twitter, Facebook, WhatsApp, Reddit, and Quora), we observe a very high bias in the dataset annotation by the human annotators~\cite{geva2019we, srivastava2020iit}. On manual inspection, we observe a significantly high ambiguity in the sentiment classification datasets for the Hinglish language (see Example I in Figure~\ref{fig:limitations}). The primary reason for this ambiguous behavior can be attributed to the underlying bias of the human annotators. 

\item \textbf{Annotator's proficiency}: The lack of grammatical standardization (see Example II \& IV in Figure~\ref{fig:limitations}) for the code-mixed languages presents the challenge of spelling variation, dialect, and readability. This standardization scarcity makes it challenging to employ human annotators for an annotation task. Also, evaluating an annotator's proficiency in a particular code-mixed language is a challenge. The natural language generation tasks such as machine translation, text generation, data to text generation, and summarization suffer heavily due to this constraint.

\item \textbf{Evaluation metrics}: The evaluation metrics developed for the monolingual languages does not seem to capture the linguistic diversity of the code-mixed languages. The majority of the widely used NLG related metrics such as Bilingual Evaluation Understudy (BLEU)~\cite{papineni2002bleu}, Word Error Rate (WER)~\cite{levenshtein1966binary}, and Translation Error Rate (TER)~\cite{snover2006study}, gives a false impression of the system performance (see Example III in Figure \ref{fig:limitations}). We need metrics that can account for the various attributes of the code-mixed languages, such as spelling variations, informal writing style, and contextual information.  

\item \textbf{Reproducibility}: Reproducibility is a major issue for the majority of the NLP research~\cite{wieling2018reproducibility}. Code-mixing, which is a relatively understudied area of research, suffers significantly due to the non-reproducible research. Be it off the self-usage or baseline comparison, the non-reproducible research is a significant bottleneck for the code-mixing research. Also, the lack of publicly available datasets limits the scope of future opportunities.

\item \textbf{Benchmark}: Recently, we witness several efforts~\cite{aguilar2020lince, khanuja2020gluecos} to benchmark various datasets and tasks in code-mixed language. The majority of these benchmarks leverage very small-scale datasets, which make generalizability and large-scale applicability nearly impossible. Most critical NLP tasks, such as question-answering, machine translation, and summarization, are still unaddressed due to either lack of research or the unavailability of publicly available datasets and systems.

\item \textbf{Official status}: Since the code-mixed languages do not possess an official status, it becomes challenging to enforce standardization. The majority of the recent efforts in NLP for the multilingual communities~\cite{kakwani2020indicnlpsuite,hu2020xtreme,xue2020mt5} do not include the code-mixed languages even though the number of speakers for such languages is very high. 
 
\end{itemize}

\begin{figure}[!tbh]
\centering
\small{
\begin{tcolorbox}[colback=white]

\begin{center} 
\hl{\textbf{Example I (Human bias)} }\\
\end{center}
\textcolor{alizarin}{\textsc{Hinglish} \cite{patwa2020semeval}}:  Twitter k baghair apna roza mumkin nahi hota ? Apna chutiyaap
dusron per thopna band karo Bhai ! \\
\textsc{\textcolor{cadmiumgreen}{Annotated sentiment}}: Positive\\
\textsc{\textcolor{blue}{Possible sentiment(s)}}: Negative, Neutral 

\begin{center} 
\hl{\textbf{Example II (Annotator's proficiency)} }\\
\end{center}

\textcolor{alizarin}{\textsc{Hinglish} \cite{patra2015shared}}: uske baap ko mobile ki factory tha kya.\\
\textsc{\textcolor{cadmiumgreen}{Incorrect annotator's language label}}: mobile- Hindi\\
\textsc{\textcolor{blue}{Correct language label}}: mobile-English

\begin{center} 
\hl{\textbf{Example III (Evaluation metrics)} }\\
\end{center}

\textcolor{alizarin}{\textsc{Source Hinglish} \cite{srivastava-singh-2020-phinc}}: kaun hai ye zaleel insaan? \\
\textsc{\textcolor{cadmiumgreen}{Reference Hinglish}}: Who is this zaleel person? \\
\textsc{\textcolor{blue}{BLEU}: 0.7598}
\textsc{\textcolor{blue}{Human score}: 1}

\begin{center} 
\hl{\textbf{Example IV (Lack of standardization)} }\\
\end{center}
\textcolor{alizarin}{\textsc{Spelling variation}} \cite{dhar2018enabling}: Jaldi $\rightarrow$ [Jldi, Jalde, Jald], Jyada $\rightarrow$ [Jada, Jyda, Jyaada] \\
\textcolor{alizarin}{\textsc{Language ambiguity}} \cite{patra2015shared}: \{to, us, is, the\} $\rightarrow$ \{English, Hindi\}

\end{tcolorbox}}
\caption{Example of the various limitations in the existing code-mixed datasets. The cited works are a representative example and not an exhaustive list.}
\label{fig:limitations}
\end{figure}

\begin{table*}[!tbh]
\resizebox{\hsize}{!}{
\small{
\begin{tabular}{|c|l|l|}
\hline
& \textbf{Possible code-mixed data source(s)} & \textbf{Already existing datasets (Language)} \\ \hline
\textbf{CDM} &\citet{shahi2020fakecovid}, \citet{li2020mm}, \citet{li2020mm},~\citet{kar2020no},~\citet{patwa2020fighting}& \citet{montesi2020understanding} (Es) \\ \hline
\textbf{CDR} &  crisisnlp.qcri.org    & \citet{imran2013extracting} (En)\\ \hline
\textbf{CDS} &  crisisnlp.qcri.org    & \citet{rudra2015extracting} (En), ~\citet{rudra2016summarizing} (En)\\ \hline
\textbf{HKG} &  patient.info, emedinexus, indiamedicalhub, healthvision.in & \citet{lindberg1993unified} (En),~\citet{oliveira2017recommendation} (En)\\ \hline
\textbf{HCB} &  nhp.gov.in, nhsrcindia.org, nathealthindia.org, ima-india.org& \citet{9190281} (En)\\ \hline
\textbf{HMI} &  thip.media, poynter.org/coronavirusfactsalliance&\citet{ZHAO2021102390} (En)\\ \hline
\textbf{PDM} &  newschecker.in, politifact.com    & \citet{wang-2017-liar} (En),~\citet{norregaard2019nela} (En) \\ \hline
\textbf{POE} & guides.nyu.edu/polisci/public-opinion-data, ucsd.libguides.com/data-statistics/publicopinion     & \citet{kannangara2019political} (En) \\ \hline
\textbf{FND} &  altnews.in, boomlive.in, smhoaxslayer.com, factchecker.in  & \citet{wang-2017-liar} (En),\citet{golbeck2018fake} (En)\\ \hline
\textbf{FNE} &  factcrescendo.com, vishvasnews.com, factly.in, maldita.es& \citet{wang-2017-liar} (En)\\ \hline
\textbf{HSD} &  hatespeechdata.com, NEWS comments \cite{hate1,hate2}& \citet{rudra2018characterizing} (En)\\ \hline
\end{tabular}}}
\caption{Various possible code-mixed data source(s) for the societal good applications. Already existing datasets in various languages could be used to generate the synthetic data for each of these tasks. Here `En' represents English and `Es' represents Spanish language. }
\label{tab:data_source}
\end{table*}

\section{The Futuristic NLP for Applications of Social Good}
\label{sec:future}

This section proposes futuristic datasets, models, and tools that can significantly advance the current state of the research in multilingual NLP applications for societal good. 

\subsection{Datasets} 
\label{sec:dataset}
As noted in the previous sections, natural code-mixed datasets are rare and small in size, with a major focus on selected applications. Here, we present some primary data sources for all the five societal good applications. We summarize our findings\footnote{Due to the ease of accessibility of the code-mixed text on the social media platforms (WhatsApp, Facebook, and Twitter), majority of the works considers these platforms. Here, we present the platforms which could also be a good alternative to curate the code-mixed datasets.} in Table \ref{tab:data_source}. There are several good quality datasets already available in various other languages for these tasks (see Table \ref{tab:data_source}, third column). We could effectively utilize these monolingual datasets to create the code-mixed datasets using various code-mixed text generation systems \cite{gupta2020semi,pratapa2018language}. Since most of these systems use parallel sentences to generate code-mixed text, we can create parallel data from state-of-the-art translation systems by first translating the original monolingual source data and then use these systems to generate the code-mixed text. We can also use this technique to convert the monolingual data from the possible code-mixed data sources (mentioned in Table \ref{tab:data_source}, second column) to get the code-mixed data. This approach can significantly address the data scarcity problem in the code-mixed languages.

\subsection{Models}
\label{sec:future_models}
The recent NLP advancements contribute to popular contextual representations constructed using (semi-)unsupervised language models~\cite{vaswani2017attention}. As these language models are trained on large monolingual datasets, they do not perform well in code-mixing tasks~\cite{khanuja2020gluecos}. We, therefore, propose the construction of pre-trained models from the large-scale code-mixed text. The proposed datasets (see Section \ref{sec:dataset}) can be used for this pretraining and further task-specific fine-tuning.

\subsection{Tools}
\label{sec:future_tools}
In the future, we envisage several NLP tools deployed in the field that can efficiently curate and process code-mixed data. For example, significant information is available to debunk a fake tweet by just looking into the replies. In most cases, the replies are comparatively more informal and contain facts in code-mixed language. The facts can be extracted and encoded into completely data-driven automatic fake news detectors for code-mixed text. This also helps in reducing political and ethical bias towards the source of fake news~\cite{van2020you}. Several media websites that curate and track hate speech are currently shutdown under political pressure. Thus, an automatic curation and tracking tool for hate speech in a multilingual community---where the text is mostly informal and contains large volumes of code-mixing---can benefit society.

\section{Conclusion}
\label{sec:conclusion}
This paper discusses five applications for social good and presents the current state of the research in each application area. We critically analyze the code-mixing research in each application area and show an extensive opportunity for future research for dataset curation, modeling, and tool development. We aim to develop and extend our expertise in handling the large volume of code-mixed data in the future. 

\clearpage

\bibliographystyle{acl_natbib}
\bibliography{naacl2021}

\begin{thebibliography}{126}
\expandafter\ifx\csname natexlab\endcsname\relax\def\natexlab#1{#1}\fi

\bibitem[{Aasman and Mirhaji(2018)}]{aasman2018knowledge}
Jans Aasman and Parsa Mirhaji. 2018.
\newblock Knowledge graph solutions in healthcare for improved clinical
  outcomes.
\newblock In \emph{CEUR Workshop Proceedings}, volume 2180. CEUR-WS.

\bibitem[{Acharoui et~al.(2020)Acharoui, Alaoui, Ettaki, Zerouaoui, and
  Dakkon}]{acharoui2020identifying}
Zakia Acharoui, Altaf Alaoui, Badia Ettaki, Jamal Zerouaoui, and Mohamed
  Dakkon. 2020.
\newblock Identifying political influencers on youtube during the 2016 moroccan
  general election.
\newblock \emph{Procedia Computer Science}, 170:1102--1109.

\bibitem[{Aguilar et~al.(2020)Aguilar, Kar, and Solorio}]{aguilar2020lince}
Gustavo Aguilar, Sudipta Kar, and Thamar Solorio. 2020.
\newblock Lince: A centralized benchmark for linguistic code-switching
  evaluation.
\newblock In \emph{Proceedings of The 12th Language Resources and Evaluation
  Conference}, pages 1803--1813.

\bibitem[{Amato et~al.(2017)Amato, Marrone, Moscato, Piantadosi, Picariello,
  and Sansone}]{amato2017chatbots}
Flora Amato, Stefano Marrone, Vincenzo Moscato, Gabriele Piantadosi, Antonio
  Picariello, and Carlo Sansone. 2017.
\newblock Chatbots meet ehealth: Automatizing healthcare.
\newblock In \emph{WAIAH@ AI* IA}, pages 40--49.

\bibitem[{Amjad et~al.(2020)Amjad, Sidorov, and Zhila}]{amjad2020data}
Maaz Amjad, Grigori Sidorov, and Alisa Zhila. 2020.
\newblock Data augmentation using machine translation for fake news detection
  in the urdu language.
\newblock In \emph{Proceedings of The 12th Language Resources and Evaluation
  Conference}, pages 2537--2542.

\bibitem[{Asghar et~al.(2014)Asghar, RahmanUllah, Khan, Ahmad, and
  Nawaz}]{asghar2014political}
Muhammad~Zubair Asghar, Ahmad~B RahmanUllah, Aurangzeb Khan, Shakeel Ahmad, and
  Irfan~Ullah Nawaz. 2014.
\newblock Political miner: opinion extraction from user generated political
  reviews.
\newblock \emph{Sci. Int (Lahore)}, 26(1):385--389.

\bibitem[{Badawy et~al.(2018)Badawy, Ferrara, and Lerman}]{badawy2018analyzing}
Adam Badawy, Emilio Ferrara, and Kristina Lerman. 2018.
\newblock Analyzing the digital traces of political manipulation: The 2016
  russian interference twitter campaign.
\newblock In \emph{2018 IEEE/ACM International Conference on Advances in Social
  Networks Analysis and Mining (ASONAM)}, pages 258--265. IEEE.

\bibitem[{Bal et~al.(2020)Bal, Sinha, Dutta, Joshi, Ghosh, and
  Dutt}]{bal2020analysing}
Rakesh Bal, Sayan Sinha, Swastika Dutta, Risabh Joshi, Sayan Ghosh, and Ritam
  Dutt. 2020.
\newblock Analysing the extent of misinformation in cancer related tweets.
\newblock In \emph{Proceedings of the International AAAI Conference on Web and
  Social Media}, volume~14, pages 924--928.

\bibitem[{Baldauf(2004)}]{baldauf2004hindi}
Scott Baldauf. 2004.
\newblock A hindi-english jumble, spoken by 350 million.
\newblock \emph{The Christian Science Monitor}, 1123(1):3.

\bibitem[{Bamman and Smith(2015)}]{bamman2015open}
David Bamman and Noah~A Smith. 2015.
\newblock Open extraction of fine-grained political statements.
\newblock In \emph{Proceedings of the 2015 Conference on Empirical Methods in
  Natural Language Processing}, pages 76--85.

\bibitem[{Barclay et~al.(2015)Barclay, Pichandy, Venkat, and
  Sudhakaran}]{barclay2015india}
Francis~P Barclay, C~Pichandy, Anusha Venkat, and Sreedevi Sudhakaran. 2015.
\newblock India 2014: Facebook ‘like’as a predictor of election outcomes.
\newblock \emph{Asian Journal of Political Science}, 23(2):134--160.

\bibitem[{Bates(2019)}]{bates2019health}
Mary Bates. 2019.
\newblock Health care chatbots are here to help.
\newblock \emph{IEEE pulse}, 10(3):12--14.

\bibitem[{Baxter and Marcella(2012)}]{baxter2012does}
Graeme Baxter and Rita Marcella. 2012.
\newblock Does scotland ‘like’this? social media use by political parties
  and candidates in scotland during the 2010 uk general election campaign.
\newblock \emph{Libri}, 62(2):109--124.

\bibitem[{Bayer and B{\'a}rd(2020)}]{hatespeech2}
Judit Bayer and Petra B{\'a}rd. 2020.
\newblock \href
  {https://www.europarl.europa.eu/thinktank/en/document.html?reference=IPOL_STU(2020)655135}
  {Hate speech and hate crime in the eu and the evaluation of online content
  regulation approaches}.

\bibitem[{Berk-Seligson(1986)}]{berk1986linguistic}
Susan Berk-Seligson. 1986.
\newblock Linguistic constraints on intrasentential code-switching: A study of
  spanish/hebrew bilingualism.
\newblock \emph{Language in society}, pages 313--348.

\bibitem[{Bibault et~al.(2019)Bibault, Chaix, Nectoux, Pienkowski,
  Guillemas{\'e}, and Brouard}]{bibault2019healthcare}
Jean-Emmanuel Bibault, Benjamin Chaix, Pierre Nectoux, Arthur Pienkowski,
  Arthur Guillemas{\'e}, and Beno{\^\i}t Brouard. 2019.
\newblock Healthcare ex machina: Are conversational agents ready for prime time
  in oncology?
\newblock \emph{Clinical and translational radiation oncology}, 16:55--59.

\bibitem[{Bohra et~al.(2018)Bohra, Vijay, Singh, Akhtar, and
  Shrivastava}]{bohra2018dataset}
Aditya Bohra, Deepanshu Vijay, Vinay Singh, Syed~Sarfaraz Akhtar, and Manish
  Shrivastava. 2018.
\newblock A dataset of hindi-english code-mixed social media text for hate
  speech detection.
\newblock In \emph{Proceedings of the second workshop on computational modeling
  of people’s opinions, personality, and emotions in social media}, pages
  36--41.

\bibitem[{Bokamba(1989)}]{bokamba1989there}
Eyamba~G Bokamba. 1989.
\newblock Are there syntactic constraints on code-mixing?
\newblock \emph{World Englishes}, 8(3):277--292.

\bibitem[{Brennen et~al.(2020)Brennen, Simon, Howard, and
  Nielsen}]{brennen2020types}
J~Scott Brennen, Felix Simon, Philip~N Howard, and Rasmus~Kleis Nielsen. 2020.
\newblock Types, sources, and claims of covid-19 misinformation.
\newblock \emph{Reuters Institute}, 7:3--1.

\bibitem[{Cameron et~al.(2018)Cameron, Cameron, Megaw, Bond, Mulvenna,
  O’Neill, Armour, and McTear}]{cameron2018best}
Gillian Cameron, David Cameron, Gavin Megaw, Raymond Bond, Maurice Mulvenna,
  Siobhan O’Neill, Cherie Armour, and Michael McTear. 2018.
\newblock Best practices for designing chatbots in mental healthcare--a case
  study on ihelpr.
\newblock In \emph{Proceedings of the 32nd International BCS Human Computer
  Interaction Conference 32}, pages 1--5.

\bibitem[{Cantarella et~al.(2020)Cantarella, Fraccaroli, and
  Volpe}]{cantarella2020does}
Michele Cantarella, Nicol{\`o} Fraccaroli, and Roberto Volpe. 2020.
\newblock Does fake news affect voting behaviour?

\bibitem[{Chaix et~al.(2019)Chaix, Bibault, Pienkowski, Delamon,
  Guillemass{\'e}, Nectoux, and Brouard}]{chaix2019chatbots}
Benjamin Chaix, Jean-Emmanuel Bibault, Arthur Pienkowski, Guillaume Delamon,
  Arthur Guillemass{\'e}, Pierre Nectoux, and Beno{\^\i}t Brouard. 2019.
\newblock When chatbots meet patients: one-year prospective study of
  conversations between patients with breast cancer and a chatbot.
\newblock \emph{Jmir Cancer}, 5(1):e12856.

\bibitem[{Chandu et~al.(2019)Chandu, Loginova, Gupta, Genabith, Neumann,
  Chinnakotla, Nyberg, and Black}]{chandu2019code}
Khyathi Chandu, Ekaterina Loginova, Vishal Gupta, Josef~van Genabith,
  G{\"u}nter Neumann, Manoj Chinnakotla, Eric Nyberg, and Alan~W Black. 2019.
\newblock Code-mixed question answering challenge: Crowd-sourcing data and
  techniques.
\newblock In \emph{Third Workshop on Computational Approaches to Linguistic
  Code-Switching}, pages 29--38. Association for Computational Linguistics
  (ACL).

\bibitem[{Choi et~al.(2017)Choi, Bahadori, Song, Stewart, and
  Sun}]{choi2017gram}
Edward Choi, Mohammad~Taha Bahadori, Le~Song, Walter~F Stewart, and Jimeng Sun.
  2017.
\newblock Gram: graph-based attention model for healthcare representation
  learning.
\newblock In \emph{Proceedings of the 23rd ACM SIGKDD International Conference
  on Knowledge Discovery and Data Mining}, pages 787--795.

\bibitem[{Chopra et~al.(2020)Chopra, Sawhney, Mathur, and
  Shah}]{chopra2020hindi}
Shivang Chopra, Ramit Sawhney, Puneet Mathur, and Rajiv~Ratn Shah. 2020.
\newblock Hindi-english hate speech detection: Author profiling, debiasing, and
  practical perspectives.
\newblock In \emph{Proceedings of the AAAI Conference on Artificial
  Intelligence}, volume~34, pages 386--393.

\bibitem[{Conversation(2016)}]{hate1}
The Conversation. 2016.
\newblock \href
  {https://theconversation.com/how-news-sites-online-comments-helped-build-our-hateful-electorate-70170}
  {Mobile messaging app map-february 2018}.
\newblock [Online; accessed 30-Jan-2021].

\bibitem[{csmonitor(2016)}]{hate2}
csmonitor. 2016.
\newblock \href
  {https://www.csmonitor.com/Technology/Breakthroughs-Voices/2016/1219/How-news-sites-online-comments-help-fuel-hatred}
  {How news sites' online comments help fuel hatred}.
\newblock [Online; accessed 30-Jan-2021].

\bibitem[{Cui and Lee(2020)}]{cui2020coaid}
Limeng Cui and Dongwon Lee. 2020.
\newblock Coaid: Covid-19 healthcare misinformation dataset.
\newblock \emph{arXiv preprint arXiv:2006.00885}.

\bibitem[{Cui et~al.(2020)Cui, Seo, Tabar, Ma, Wang, and
  Lee}]{cui2020deterrent}
Limeng Cui, Haeseung Seo, Maryam Tabar, Fenglong Ma, Suhang Wang, and Dongwon
  Lee. 2020.
\newblock Deterrent: Knowledge guided graph attention network for detecting
  healthcare misinformation.
\newblock In \emph{Proceedings of the 26th ACM SIGKDD International Conference
  on Knowledge Discovery \& Data Mining}, pages 492--502.

\bibitem[{Das and Gamb{\"a}ck(2015)}]{das2015code}
Amitava Das and Bj{\"o}rn Gamb{\"a}ck. 2015.
\newblock Code-mixing in social media text: the last language identification
  frontier?

\bibitem[{Davidson et~al.(2017)Davidson, Warmsley, Macy, and
  Weber}]{davidson2017automated}
Thomas Davidson, Dana Warmsley, Michael Macy, and Ingmar Weber. 2017.
\newblock Automated hate speech detection and the problem of offensive
  language.
\newblock In \emph{Proceedings of the International AAAI Conference on Web and
  Social Media}, volume~11.

\bibitem[{Dhar et~al.(2018)Dhar, Kumar, and Shrivastava}]{dhar2018enabling}
Mrinal Dhar, Vaibhav Kumar, and Manish Shrivastava. 2018.
\newblock Enabling code-mixed translation: Parallel corpus creation and mt
  augmentation approach.
\newblock In \emph{Proceedings of the First Workshop on Linguistic Resources
  for Natural Language Processing}, pages 131--140.

\bibitem[{Divya et~al.(2018)Divya, Indumathi, Ishwarya, Priyasankari, and
  Devi}]{divya2018self}
S~Divya, V~Indumathi, S~Ishwarya, M~Priyasankari, and S~Kalpana Devi. 2018.
\newblock A self-diagnosis medical chatbot using artificial intelligence.
\newblock \emph{Journal of Web Development and Web Designing}, 3(1):1--7.

\bibitem[{Futurist(2020)}]{healthcare}
The~Medical Futurist. 2020.
\newblock \href {https://medicalfuturist.com/top-12-health-chatbots/} {The top
  12 health chatbots}.

\bibitem[{Garimella and Eckles(2020)}]{garimella2020images}
Kiran Garimella and Dean Eckles. 2020.
\newblock Images and misinformation in political groups: Evidence from whatsapp
  in india.
\newblock \emph{arXiv preprint arXiv:2005.09784}.

\bibitem[{Geva et~al.(2019)Geva, Goldberg, and Berant}]{geva2019we}
Mor Geva, Yoav Goldberg, and Jonathan Berant. 2019.
\newblock Are we modeling the task or the annotator? an investigation of
  annotator bias in natural language understanding datasets.
\newblock In \emph{Proceedings of the 2019 Conference on Empirical Methods in
  Natural Language Processing and the 9th International Joint Conference on
  Natural Language Processing (EMNLP-IJCNLP)}, pages 1161--1166.

\bibitem[{Golbeck et~al.(2018)Golbeck, Mauriello, Auxier, Bhanushali, Bonk,
  Bouzaghrane, Buntain, Chanduka, Cheakalos, Everett et~al.}]{golbeck2018fake}
Jennifer Golbeck, Matthew Mauriello, Brooke Auxier, Keval~H Bhanushali,
  Christopher Bonk, Mohamed~Amine Bouzaghrane, Cody Buntain, Riya Chanduka,
  Paul Cheakalos, Jennine~B Everett, et~al. 2018.
\newblock Fake news vs satire: A dataset and analysis.
\newblock In \emph{Proceedings of the 10th ACM Conference on Web Science},
  pages 17--21.

\bibitem[{Grinberg et~al.(2019)Grinberg, Joseph, Friedland, Swire-Thompson, and
  Lazer}]{grinberg2019fake}
Nir Grinberg, Kenneth Joseph, Lisa Friedland, Briony Swire-Thompson, and David
  Lazer. 2019.
\newblock Fake news on twitter during the 2016 us presidential election.
\newblock \emph{Science}, 363(6425):374--378.

\bibitem[{Guess et~al.(2019)Guess, Nagler, and Tucker}]{guess2019less}
Andrew Guess, Jonathan Nagler, and Joshua Tucker. 2019.
\newblock Less than you think: Prevalence and predictors of fake news
  dissemination on facebook.
\newblock \emph{Science advances}, 5(1):eaau4586.

\bibitem[{Gull et~al.(2016)Gull, Shoaib, Rasheed, Abid, and
  Zahoor}]{gull2016pre}
Ratab Gull, Umar Shoaib, Saba Rasheed, Washma Abid, and Beenish Zahoor. 2016.
\newblock Pre processing of twitter's data for opinion mining in political
  context.
\newblock \emph{Procedia Computer Science}, 96:1560--1570.

\bibitem[{Gumperz and Hernandez(1969)}]{gumperz1969cognitive}
John~Joseph Gumperz and Edward Hernandez. 1969.
\newblock \emph{Cognitive aspects of bilingual communication}.
\newblock 28. Institute of International Studies, University of California.

\bibitem[{Gupta et~al.(2020)Gupta, Ekbal, and Bhattacharyya}]{gupta2020semi}
Deepak Gupta, Asif Ekbal, and Pushpak Bhattacharyya. 2020.
\newblock A semi-supervised approach to generate the code-mixed text using
  pre-trained encoder and transfer learning.
\newblock In \emph{Proceedings of the 2020 Conference on Empirical Methods in
  Natural Language Processing: Findings}, pages 2267--2280.

\bibitem[{Hassan et~al.(2015)Hassan, Adair, Hamilton, Li, Tremayne, Yang, and
  Yu}]{hassan2015quest}
Naeemul Hassan, Bill Adair, James~T Hamilton, Chengkai Li, Mark Tremayne, Jun
  Yang, and Cong Yu. 2015.
\newblock The quest to automate fact-checking.
\newblock In \emph{Proceedings of the 2015 Computation+ Journalism Symposium}.

\bibitem[{Hassan et~al.(2017)Hassan, Zhang, Arslan, Caraballo, Jimenez,
  Gawsane, Hasan, Joseph, Kulkarni, Nayak et~al.}]{hassan2017claimbuster}
Naeemul Hassan, Gensheng Zhang, Fatma Arslan, Josue Caraballo, Damian Jimenez,
  Siddhant Gawsane, Shohedul Hasan, Minumol Joseph, Aaditya Kulkarni,
  Anil~Kumar Nayak, et~al. 2017.
\newblock Claimbuster: the first-ever end-to-end fact-checking system.
\newblock \emph{Proceedings of the VLDB Endowment}, 10(12):1945--1948.

\bibitem[{Heredia et~al.(2018)Heredia, Prusa, and
  Khoshgoftaar}]{heredia2018impact}
Brian Heredia, Joseph~D Prusa, and Taghi~M Khoshgoftaar. 2018.
\newblock The impact of malicious accounts on political tweet sentiment.
\newblock In \emph{2018 IEEE 4th International Conference on Collaboration and
  Internet Computing (CIC)}, pages 197--202. IEEE.

\bibitem[{Hossain et~al.(2020)Hossain, Rahman, Islam, and
  Kar}]{hossain2020banfakenews}
Md~Zobaer Hossain, Md~Ashraful Rahman, Md~Saiful Islam, and Sudipta Kar. 2020.
\newblock Banfakenews: A dataset for detecting fake news in bangla.
\newblock In \emph{Proceedings of The 12th Language Resources and Evaluation
  Conference}, pages 2862--2871.

\bibitem[{Hou et~al.(2019)Hou, P{\'e}rez-Rosas, Loeb, and
  Mihalcea}]{hou2019towards}
Rui Hou, Ver{\'o}nica P{\'e}rez-Rosas, Stacy Loeb, and Rada Mihalcea. 2019.
\newblock Towards automatic detection of misinformation in online medical
  videos.
\newblock In \emph{2019 International conference on multimodal interaction},
  pages 235--243.

\bibitem[{Hu et~al.(2020)Hu, Ruder, Siddhant, Neubig, Firat, and
  Johnson}]{hu2020xtreme}
Junjie Hu, Sebastian Ruder, Aditya Siddhant, Graham Neubig, Orhan Firat, and
  Melvin Johnson. 2020.
\newblock Xtreme: A massively multilingual multi-task benchmark for evaluating
  cross-lingual generalisation.
\newblock In \emph{International Conference on Machine Learning}, pages
  4411--4421. PMLR.

\bibitem[{Imran et~al.(2013)Imran, Elbassuoni, Castillo, Diaz, and
  Meier}]{imran2013extracting}
Muhammad Imran, Shady Elbassuoni, Carlos Castillo, Fernando Diaz, and Patrick
  Meier. 2013.
\newblock Extracting information nuggets from disaster-related messages in
  social media.
\newblock In \emph{Iscram}.

\bibitem[{Joshi et~al.(2016)Joshi, Bhattacharyya, and
  Carman}]{joshi2016political}
Aditya Joshi, Pushpak Bhattacharyya, and Mark Carman. 2016.
\newblock Political issue extraction model: A novel hierarchical topic model
  that uses tweets by political and non-political authors.
\newblock In \emph{Proceedings of the 7th Workshop on Computational Approaches
  to Subjectivity, Sentiment and Social Media Analysis}, pages 82--90.

\bibitem[{Kakwani et~al.(2020)Kakwani, Kunchukuttan, Golla, N.C.,
  Bhattacharyya, Khapra, and Kumar}]{kakwani2020indicnlpsuite}
Divyanshu Kakwani, Anoop Kunchukuttan, Satish Golla, Gokul N.C., Avik
  Bhattacharyya, Mitesh~M. Khapra, and Pratyush Kumar. 2020.
\newblock {IndicNLPSuite: Monolingual Corpora, Evaluation Benchmarks and
  Pre-trained Multilingual Language Models for Indian Languages}.
\newblock In \emph{Findings of EMNLP}.

\bibitem[{Kamble and Joshi(2018)}]{kamble2018hate}
Satyajit Kamble and Aditya Joshi. 2018.
\newblock Hate speech detection from code-mixed hindi-english tweets using deep
  learning models.
\newblock In \emph{15th International Conference on Natural Language
  Processing}, page 155.

\bibitem[{Kandias et~al.(2013)Kandias, Mitrou, Stavrou, and
  Gritzalis}]{kandias2013youtube}
Miltiadis Kandias, Lilian Mitrou, Vasilis Stavrou, and Dimitris Gritzalis.
  2013.
\newblock Youtube user and usage profiling: Stories of political horror and
  security success.
\newblock In \emph{International Conference on E-Business and
  Telecommunications}, pages 270--289. Springer.

\bibitem[{Kannangara(2019)}]{kannangara2019political}
Sandeepa~Harshanganie Kannangara. 2019.
\newblock Political opinion mining and analysis for social media.

\bibitem[{Kanungo(2015)}]{kanungo2015india}
Neena~Talwar Kanungo. 2015.
\newblock India’s digital poll battle: Political parties and social media in
  the 16th lok sabha elections.
\newblock \emph{Studies in Indian Politics}, 3(2):212--228.

\bibitem[{Kar et~al.(2020)Kar, Bhardwaj, Samanta, and Azad}]{kar2020no}
Debanjana Kar, Mohit Bhardwaj, Suranjana Samanta, and Amar~Prakash Azad. 2020.
\newblock No rumours please! a multi-indic-lingual approach for covid
  fake-tweet detection.
\newblock \emph{arXiv preprint arXiv:2010.06906}.

\bibitem[{Kaschesky et~al.(2011)Kaschesky, Sobkowicz, and
  Bouchard}]{kaschesky2011opinion}
Michael Kaschesky, Pawel Sobkowicz, and Guillaume Bouchard. 2011.
\newblock Opinion mining in social media: modeling, simulating, and visualizing
  political opinion formation in the web.
\newblock In \emph{Proceedings of the 12th Annual International Digital
  Government Research Conference: Digital Government Innovation in Challenging
  Times}, pages 317--326.

\bibitem[{Khaldarova and Pantti(2016)}]{doi:10.1080/17512786.2016.1163237}
Irina Khaldarova and Mervi Pantti. 2016.
\newblock \href {https://doi.org/10.1080/17512786.2016.1163237} {Fake news}.
\newblock \emph{Journalism Practice}, 10(7):891--901.

\bibitem[{Khanuja et~al.(2020)Khanuja, Dandapat, Srinivasan, Sitaram, and
  Choudhury}]{khanuja2020gluecos}
Simran Khanuja, Sandipan Dandapat, Anirudh Srinivasan, Sunayana Sitaram, and
  Monojit Choudhury. 2020.
\newblock Gluecos: An evaluation benchmark for code-switched nlp.
\newblock In \emph{Proceedings of the 58th Annual Meeting of the Association
  for Computational Linguistics}, pages 3575--3585.

\bibitem[{Kouzy et~al.(2020)Kouzy, Abi~Jaoude, Kraitem, El~Alam, Karam, Adib,
  Zarka, Traboulsi, Akl, and Baddour}]{kouzy2020coronavirus}
Ramez Kouzy, Joseph Abi~Jaoude, Afif Kraitem, Molly~B El~Alam, Basil Karam,
  Elio Adib, Jabra Zarka, Cindy Traboulsi, Elie~W Akl, and Khalil Baddour.
  2020.
\newblock Coronavirus goes viral: quantifying the covid-19 misinformation
  epidemic on twitter.
\newblock \emph{Cureus}, 12(3).

\bibitem[{Kowatsch et~al.(2017)Kowatsch, Ni{\ss}en, Shih, R{\"u}egger, Volland,
  Filler, K{\"u}nzler, Barata, B{\"u}chter, Brogle et~al.}]{kowatsch2017text}
Tobias Kowatsch, Marcia Ni{\ss}en, Chen-Hsuan~Iris Shih, Dominik R{\"u}egger,
  Dirk Volland, Andreas Filler, Florian K{\"u}nzler, Filipe Barata, Dirk
  B{\"u}chter, Bj{\"o}rn Brogle, et~al. 2017.
\newblock Text-based healthcare chatbots supporting patient and health
  professional teams: preliminary results of a randomized controlled trial on
  childhood obesity.

\bibitem[{Kumar et~al.(2018)Kumar, Reganti, Bhatia, and
  Maheshwari}]{kumar2018aggression}
Ritesh Kumar, Aishwarya~N Reganti, Akshit Bhatia, and Tushar Maheshwari. 2018.
\newblock Aggression-annotated corpus of hindi-english code-mixed data.
\newblock In \emph{Proceedings of the Eleventh International Conference on
  Language Resources and Evaluation (LREC 2018)}.

\bibitem[{Kwanda and Lin(2020)}]{kwanda2020fake}
Febbie~Austina Kwanda and Trisha~TC Lin. 2020.
\newblock Fake news practices in indonesian newsrooms during and after the palu
  earthquake: a hierarchy-of-influences approach.
\newblock \emph{Information, Communication \& Society}, pages 1--18.

\bibitem[{Levenshtein(1966)}]{levenshtein1966binary}
Vladimir~I Levenshtein. 1966.
\newblock Binary codes capable of correcting deletions, insertions, and
  reversals.
\newblock In \emph{Soviet physics doklady}, volume~10, pages 707--710.

\bibitem[{Li et~al.(2020)Li, Jiang, Shu, and Liu}]{li2020mm}
Yichuan Li, Bohan Jiang, Kai Shu, and Huan Liu. 2020.
\newblock Mm-covid: A multilingual and multidimensional data repository for
  combating covid-19 fake news.
\newblock \emph{arXiv preprint arXiv:2011.04088}.

\bibitem[{Lindberg et~al.(1993)Lindberg, Humphreys, and
  McCray}]{lindberg1993unified}
Donald~AB Lindberg, Betsy~L Humphreys, and Alexa~T McCray. 1993.
\newblock The unified medical language system.
\newblock \emph{Methods of information in medicine}, 32(4):281.

\bibitem[{van~der Linden et~al.(2020)van~der Linden, Panagopoulos, and
  Roozenbeek}]{van2020you}
Sander van~der Linden, Costas Panagopoulos, and Jon Roozenbeek. 2020.
\newblock You are fake news: political bias in perceptions of fake news.
\newblock \emph{Media, Culture \& Society}, 42(3):460--470.

\bibitem[{Ma et~al.(2018)Ma, You, Xiao, Chitta, Zhou, and Gao}]{ma2018kame}
Fenglong Ma, Quanzeng You, Houping Xiao, Radha Chitta, Jing Zhou, and Jing Gao.
  2018.
\newblock Kame: Knowledge-based attention model for diagnosis prediction in
  healthcare.
\newblock In \emph{Proceedings of the 27th ACM International Conference on
  Information and Knowledge Management}, pages 743--752.

\bibitem[{Machado et~al.(2019)Machado, Kira, Narayanan, Kollanyi, and
  Howard}]{machado2019study}
Caio Machado, Beatriz Kira, Vidya Narayanan, Bence Kollanyi, and Philip Howard.
  2019.
\newblock A study of misinformation in whatsapp groups with a focus on the
  brazilian presidential elections.
\newblock In \emph{Companion proceedings of the 2019 World Wide Web
  conference}, pages 1013--1019.

\bibitem[{Mandl et~al.(2019)Mandl, Modha, Majumder, Patel, Dave, Mandlia, and
  Patel}]{mandl2019overview}
Thomas Mandl, Sandip Modha, Prasenjit Majumder, Daksh Patel, Mohana Dave,
  Chintak Mandlia, and Aditya Patel. 2019.
\newblock Overview of the hasoc track at fire 2019: Hate speech and offensive
  content identification in indo-european languages.
\newblock In \emph{Proceedings of the 11th Forum for Information Retrieval
  Evaluation}, pages 14--17.

\bibitem[{Mathew et~al.(2020)Mathew, Saha, Yimam, Biemann, Goyal, and
  Mukherjee}]{mathew2020hatexplain}
Binny Mathew, Punyajoy Saha, Seid~Muhie Yimam, Chris Biemann, Pawan Goyal, and
  Animesh Mukherjee. 2020.
\newblock Hatexplain: A benchmark dataset for explainable hate speech
  detection.
\newblock \emph{arXiv preprint arXiv:2012.10289}.

\bibitem[{Meel and Vishwakarma(2019)}]{meel2019fake}
Priyanka Meel and Dinesh~Kumar Vishwakarma. 2019.
\newblock Fake news, rumor, information pollution in social media and web: A
  contemporary survey of state-of-the-arts, challenges and opportunities.
\newblock \emph{Expert Systems with Applications}, page 112986.

\bibitem[{Mena(2020)}]{mena2020cleaning}
Paul Mena. 2020.
\newblock Cleaning up social media: The effect of warning labels on likelihood
  of sharing false news on facebook.
\newblock \emph{Policy \& internet}, 12(2):165--183.

\bibitem[{Montesi(2020)}]{montesi2020understanding}
Michela Montesi. 2020.
\newblock Understanding fake news during the covid-19 health crisis from the
  perspective of information behaviour: The case of spain.
\newblock \emph{Journal of Librarianship and Information Science}, page
  0961000620949653.

\bibitem[{Myslinski(2015)}]{myslinski2015method}
Lucas~J Myslinski. 2015.
\newblock Method of and system for validating a fact checking system.
\newblock US Patent 9,087,048.

\bibitem[{Nakamura et~al.(2020)Nakamura, Levy, and Wang}]{nakamura2020fakeddit}
Kai Nakamura, Sharon Levy, and William~Yang Wang. 2020.
\newblock Fakeddit: A new multimodal benchmark dataset for fine-grained fake
  news detection.
\newblock In \emph{Proceedings of The 12th Language Resources and Evaluation
  Conference}, pages 6149--6157.

\bibitem[{Neudert et~al.(2017)Neudert, Kollanyi, and Howard}]{neudert2017junk}
L~Neudert, Bence Kollanyi, and Philip~N Howard. 2017.
\newblock Junk news and bots during the german parliamentary election: What are
  german voters sharing over twitter?

\bibitem[{Nockleby(2000)}]{nockleby2000hate}
John~T Nockleby. 2000.
\newblock Hate speech.
\newblock \emph{Encyclopedia of the American constitution}, 3(2):1277--1279.

\bibitem[{N{\o}rregaard et~al.(2019)N{\o}rregaard, Horne, and
  Adal{\i}}]{norregaard2019nela}
Jeppe N{\o}rregaard, Benjamin~D Horne, and Sibel Adal{\i}. 2019.
\newblock Nela-gt-2018: A large multi-labelled news dataset for the study of
  misinformation in news articles.
\newblock In \emph{Proceedings of the International AAAI Conference on Web and
  Social Media}, volume~13, pages 630--638.

\bibitem[{Oliveira et~al.(2017)Oliveira, Delgado, and
  Assaife}]{oliveira2017recommendation}
Jonice Oliveira, Carla Delgado, and Ana~Carolina Assaife. 2017.
\newblock A recommendation approach for consuming linked open data.
\newblock \emph{Expert Systems with Applications}, 72:407--420.

\bibitem[{Oshikawa et~al.(2020)Oshikawa, Qian, and Wang}]{oshikawa2020survey}
Ray Oshikawa, Jing Qian, and William~Yang Wang. 2020.
\newblock A survey on natural language processing for fake news detection.
\newblock In \emph{Proceedings of The 12th Language Resources and Evaluation
  Conference}, pages 6086--6093.

\bibitem[{Palanica et~al.(2019)Palanica, Flaschner, Thommandram, Li, and
  Fossat}]{palanica2019physicians}
Adam Palanica, Peter Flaschner, Anirudh Thommandram, Michael Li, and Yan
  Fossat. 2019.
\newblock Physicians’ perceptions of chatbots in health care: cross-sectional
  web-based survey.
\newblock \emph{Journal of medical Internet research}, 21(4):e12887.

\bibitem[{Papineni et~al.(2002)Papineni, Roukos, Ward, and
  Zhu}]{papineni2002bleu}
Kishore Papineni, Salim Roukos, Todd Ward, and Wei-Jing Zhu. 2002.
\newblock Bleu: a method for automatic evaluation of machine translation.
\newblock In \emph{Proceedings of the 40th annual meeting on association for
  computational linguistics}, pages 311--318. Association for Computational
  Linguistics.

\bibitem[{Patra et~al.(2015)Patra, Das, Das, and Prasath}]{patra2015shared}
Braja~Gopal Patra, Dipankar Das, Amitava Das, and Rajendra Prasath. 2015.
\newblock Shared task on sentiment analysis in indian languages (sail)
  tweets-an overview.
\newblock In \emph{International Conference on Mining Intelligence and
  Knowledge Exploration}, pages 650--655. Springer.

\bibitem[{Patwa et~al.(2020{\natexlab{a}})Patwa, Aguilar, Kar, Pandey,
  Srinivas, Gamb{\"a}ck, Chakraborty, Solorio, and Das}]{patwa2020semeval}
Parth Patwa, Gustavo Aguilar, Sudipta Kar, Suraj Pandey, PYKL Srinivas,
  Bj{\"o}rn Gamb{\"a}ck, Tanmoy Chakraborty, Thamar Solorio, and Amitava Das.
  2020{\natexlab{a}}.
\newblock Semeval-2020 task 9: Overview of sentiment analysis of code-mixed
  tweets.
\newblock In \emph{Proceedings of the Fourteenth Workshop on Semantic
  Evaluation}, pages 774--790.

\bibitem[{Patwa et~al.(2020{\natexlab{b}})Patwa, Sharma, PYKL, Guptha, Kumari,
  Akhtar, Ekbal, Das, and Chakraborty}]{patwa2020fighting}
Parth Patwa, Shivam Sharma, Srinivas PYKL, Vineeth Guptha, Gitanjali Kumari,
  Md~Shad Akhtar, Asif Ekbal, Amitava Das, and Tanmoy Chakraborty.
  2020{\natexlab{b}}.
\newblock Fighting an infodemic: Covid-19 fake news dataset.
\newblock \emph{arXiv preprint arXiv:2011.03327}.

\bibitem[{Pennycook et~al.(2020)Pennycook, McPhetres, Zhang, Lu, and
  Rand}]{pennycook2020fighting}
Gordon Pennycook, Jonathon McPhetres, Yunhao Zhang, Jackson~G Lu, and David~G
  Rand. 2020.
\newblock Fighting covid-19 misinformation on social media: Experimental
  evidence for a scalable accuracy-nudge intervention.
\newblock \emph{Psychological science}, 31(7):770--780.

\bibitem[{Phillips and Young(2009)}]{phillips2009online}
David Phillips and Philip Young. 2009.
\newblock \emph{Online public relations: A practical guide to developing an
  online strategy in the world of social media}.
\newblock Kogan Page Publishers.

\bibitem[{POPLACK(1980)}]{poplack1980sometimes}
S~POPLACK. 1980.
\newblock Sometimes i'll start a sentence in spanish y termino en espanol:
  Toward a typology of code-switching.
\newblock \emph{Linguistics. An Interdisciplinary Journal of the Language
  Sciences La Haye}, 18(7-8):581--618.

\bibitem[{Prasad and Francescutti(2017)}]{PRASAD2017215}
Abhaya~S. Prasad and Louis~Hugo Francescutti. 2017.
\newblock \href
  {https://doi.org/https://doi.org/10.1016/B978-0-12-803678-5.00519-1} {Natural
  disasters}.
\newblock In Stella~R. Quah, editor, \emph{International Encyclopedia of Public
  Health (Second Edition)}, second edition edition, pages 215 -- 222. Academic
  Press, Oxford.

\bibitem[{Pratapa et~al.(2018)Pratapa, Bhat, Choudhury, Sitaram, Dandapat, and
  Bali}]{pratapa2018language}
Adithya Pratapa, Gayatri Bhat, Monojit Choudhury, Sunayana Sitaram, Sandipan
  Dandapat, and Kalika Bali. 2018.
\newblock Language modeling for code-mixing: The role of linguistic theory
  based synthetic data.
\newblock In \emph{Proceedings of the 56th Annual Meeting of the Association
  for Computational Linguistics (Volume 1: Long Papers)}, pages 1543--1553.

\bibitem[{Qazi et~al.(2020)Qazi, Imran, and Ofli}]{qazi2020geocov19}
Umair Qazi, Muhammad Imran, and Ferda Ofli. 2020.
\newblock Geocov19: a dataset of hundreds of millions of multilingual covid-19
  tweets with location information.
\newblock \emph{SIGSPATIAL Special}, 12(1):6--15.

\bibitem[{Rashkin et~al.(2017)Rashkin, Choi, Jang, Volkova, and
  Choi}]{rashkin2017truth}
Hannah Rashkin, Eunsol Choi, Jin~Yea Jang, Svitlana Volkova, and Yejin Choi.
  2017.
\newblock Truth of varying shades: Analyzing language in fake news and
  political fact-checking.
\newblock In \emph{Proceedings of the 2017 conference on empirical methods in
  natural language processing}, pages 2931--2937.

\bibitem[{Reis et~al.(2019)Reis, Correia, Murai, Veloso, and
  Benevenuto}]{reis2019explainable}
Julio~CS Reis, Andr{\'e} Correia, Fabr{\'\i}cio Murai, Adriano Veloso, and
  Fabr{\'\i}cio Benevenuto. 2019.
\newblock Explainable machine learning for fake news detection.
\newblock In \emph{Proceedings of the 10th ACM Conference on Web Science},
  pages 17--26.

\bibitem[{Resende et~al.(2019)Resende, Melo, CS~Reis, Vasconcelos, Almeida, and
  Benevenuto}]{resende2019analyzing}
Gustavo Resende, Philipe Melo, Julio CS~Reis, Marisa Vasconcelos, Jussara~M
  Almeida, and Fabr{\'\i}cio Benevenuto. 2019.
\newblock Analyzing textual (mis) information shared in whatsapp groups.
\newblock In \emph{Proceedings of the 10th ACM Conference on Web Science},
  pages 225--234.

\bibitem[{Rudra et~al.(2016)Rudra, Banerjee, Ganguly, Goyal, Imran, and
  Mitra}]{rudra2016summarizing}
Koustav Rudra, Siddhartha Banerjee, Niloy Ganguly, Pawan Goyal, Muhammad Imran,
  and Prasenjit Mitra. 2016.
\newblock Summarizing situational tweets in crisis scenario.
\newblock In \emph{Proceedings of the 27th ACM Conference on Hypertext and
  Social Media}, pages 137--147.

\bibitem[{Rudra et~al.(2015)Rudra, Ghosh, Ganguly, Goyal, and
  Ghosh}]{rudra2015extracting}
Koustav Rudra, Subham Ghosh, Niloy Ganguly, Pawan Goyal, and Saptarshi Ghosh.
  2015.
\newblock Extracting situational information from microblogs during disaster
  events: a classification-summarization approach.
\newblock In \emph{Proceedings of the 24th ACM International on Conference on
  Information and Knowledge Management}, pages 583--592.

\bibitem[{Rudra et~al.(2018{\natexlab{a}})Rudra, Goyal, Ganguly, Mitra, and
  Imran}]{rudra2018identifying}
Koustav Rudra, Pawan Goyal, Niloy Ganguly, Prasenjit Mitra, and Muhammad Imran.
  2018{\natexlab{a}}.
\newblock Identifying sub-events and summarizing disaster-related information
  from microblogs.
\newblock In \emph{The 41st International ACM SIGIR Conference on Research \&
  Development in Information Retrieval}, pages 265--274.

\bibitem[{Rudra et~al.(2018{\natexlab{b}})Rudra, Sharma, Ganguly, and
  Ghosh}]{rudra2018characterizing}
Koustav Rudra, Ashish Sharma, Niloy Ganguly, and Saptarshi Ghosh.
  2018{\natexlab{b}}.
\newblock Characterizing and countering communal microblogs during disaster
  events.
\newblock \emph{IEEE Transactions on Computational Social Systems},
  5(2):403--417.

\bibitem[{Sadeghi and Lehmann(2019)}]{sadeghi2019linking}
Afshin Sadeghi and Jens Lehmann. 2019.
\newblock Linking physicians to medical research results via knowledge graph
  embeddings and twitter.
\newblock In \emph{Joint European Conference on Machine Learning and Knowledge
  Discovery in Databases}, pages 622--630. Springer.

\bibitem[{Salem et~al.(2019)Salem, Al~Feel, Elbassuoni, Jaber, and
  Farah}]{salem2019fa}
Fatima K~Abu Salem, Roaa Al~Feel, Shady Elbassuoni, Mohamad Jaber, and May
  Farah. 2019.
\newblock Fa-kes: A fake news dataset around the syrian war.
\newblock In \emph{Proceedings of the International AAAI Conference on Web and
  Social Media}, volume~13, pages 573--582.

\bibitem[{Santosh and Aravind(2019)}]{santosh2019hate}
TYSS Santosh and KVS Aravind. 2019.
\newblock Hate speech detection in hindi-english code-mixed social media text.
\newblock In \emph{Proceedings of the ACM India Joint International Conference
  on Data Science and Management of Data}, pages 310--313.

\bibitem[{Schmidt and Wiegand(2017)}]{schmidt2017survey}
Anna Schmidt and Michael Wiegand. 2017.
\newblock A survey on hate speech detection using natural language processing.
\newblock In \emph{Proceedings of the Fifth International workshop on natural
  language processing for social media}, pages 1--10.

\bibitem[{Schwarz et~al.(2020)Schwarz, The{\'o}philo, and
  Rocha}]{schwarz2020emet}
Stephane Schwarz, Ant{\^o}nio The{\'o}philo, and Anderson Rocha. 2020.
\newblock Emet: Embeddings from multilingual-encoder transformer for fake news
  detection.
\newblock In \emph{ICASSP 2020-2020 IEEE International Conference on Acoustics,
  Speech and Signal Processing (ICASSP)}, pages 2777--2781. IEEE.

\bibitem[{Shahi and Nandini(2020)}]{shahi2020fakecovid}
Gautam~Kishore Shahi and Durgesh Nandini. 2020.
\newblock Fakecovid--a multilingual cross-domain fact check news dataset for
  covid-19.
\newblock \emph{arXiv preprint arXiv:2006.11343}.

\bibitem[{Shin et~al.(2017)Shin, Jian, Driscoll, and Bar}]{shin2017political}
Jieun Shin, Lian Jian, Kevin Driscoll, and Fran{\c{c}}ois Bar. 2017.
\newblock Political rumoring on twitter during the 2012 us presidential
  election: Rumor diffusion and correction.
\newblock \emph{new media \& society}, 19(8):1214--1235.

\bibitem[{Shu et~al.(2019)Shu, Cui, Wang, Lee, and Liu}]{shu2019defend}
Kai Shu, Limeng Cui, Suhang Wang, Dongwon Lee, and Huan Liu. 2019.
\newblock defend: Explainable fake news detection.
\newblock In \emph{Proceedings of the 25th ACM SIGKDD International Conference
  on Knowledge Discovery \& Data Mining}, pages 395--405.

\bibitem[{Shu et~al.(2017)Shu, Sliva, Wang, Tang, and Liu}]{shu2017fake}
Kai Shu, Amy Sliva, Suhang Wang, Jiliang Tang, and Huan Liu. 2017.
\newblock Fake news detection on social media: A data mining perspective.
\newblock \emph{ACM SIGKDD explorations newsletter}, 19(1):22--36.

\bibitem[{Singh(2020)}]{hatespeech1}
Vijaita Singh. 2020.
\newblock \href
  {https://www.thehindu.com/news/national/centre-moves-for-law-on-online-abuse/article23295440.ece}
  {Centre plans law on online hate speech}.

\bibitem[{Snover et~al.(2006)Snover, Dorr, Schwartz, Micciulla, and
  Makhoul}]{snover2006study}
Matthew Snover, Bonnie Dorr, Richard Schwartz, Linnea Micciulla, and John
  Makhoul. 2006.
\newblock A study of translation edit rate with targeted human annotation.
\newblock In \emph{Proceedings of association for machine translation in the
  Americas}, volume 200. Cambridge, MA.

\bibitem[{Srivastava and Singh(2020{\natexlab{a}})}]{srivastava2020iit}
Vivek Srivastava and Mayank Singh. 2020{\natexlab{a}}.
\newblock Iit gandhinagar at semeval-2020 task 9: Code-mixed sentiment
  classification using candidate sentence generation and selection.
\newblock In \emph{Proceedings of the Fourteenth Workshop on Semantic
  Evaluation}, pages 1259--1264.

\bibitem[{Srivastava and
  Singh(2020{\natexlab{b}})}]{srivastava-singh-2020-phinc}
Vivek Srivastava and Mayank Singh. 2020{\natexlab{b}}.
\newblock \href {https://doi.org/10.18653/v1/2020.wnut-1.7} {{PHINC}: A
  parallel {H}inglish social media code-mixed corpus for machine translation}.
\newblock In \emph{Proceedings of the Sixth Workshop on Noisy User-generated
  Text (W-NUT 2020)}, pages 41--49, Online. Association for Computational
  Linguistics.

\bibitem[{Sujeet K~Sharma and Chandwani(2020)}]{fakeonline2}
Jang Bahadur~Singh Sujeet K~Sharma and Rajesh Chandwani. 2020.
\newblock \href
  {https://government.economictimes.indiatimes.com/news/digital-india/rumors-vs-fake-news-how-to-address-misinformation-in-crisis/76421449}
  {Rumors vs fake news: How to address misinformation in crisis?}

\bibitem[{Tasnim et~al.(2020)Tasnim, Hossain, and Mazumder}]{tasnim2020impact}
Samia Tasnim, Md~Mahbub Hossain, and Hoimonty Mazumder. 2020.
\newblock Impact of rumors and misinformation on covid-19 in social media.
\newblock \emph{Journal of preventive medicine and public health},
  53(3):171--174.

\bibitem[{Thorne and Vlachos(2018)}]{thorne2018automated}
James Thorne and Andreas Vlachos. 2018.
\newblock Automated fact checking: Task formulations, methods and future
  directions.
\newblock In \emph{Proceedings of the 27th International Conference on
  Computational Linguistics}, pages 3346--3359.

\bibitem[{{Ur Rahman Khilji} et~al.(2020){Ur Rahman Khilji}, {Laskar},
  {Pakray}, {Kadir}, {Lydia}, and {Bandyopadhyay}}]{9190281}
A.~F. {Ur Rahman Khilji}, S.~R. {Laskar}, P.~{Pakray}, R.~A. {Kadir}, M.~S.
  {Lydia}, and S.~{Bandyopadhyay}. 2020.
\newblock \href {https://doi.org/10.1109/DATABIA50434.2020.9190281} {Healfavor:
  Dataset and a prototype system for healthcare chatbot}.
\newblock In \emph{2020 International Conference on Data Science, Artificial
  Intelligence, and Business Analytics (DATABIA)}, pages 1--4.

\bibitem[{Vald{\'e}s(2007)}]{valdes2007multilingualism}
G~Vald{\'e}s. 2007.
\newblock Multilingualism. linguistic society of america.
\newblock \emph{On WWW at http://www. lsadc. org/info/ling-fields-multi. cfm.
  Accessed}, 9.

\bibitem[{Vaswani et~al.(2017)Vaswani, Shazeer, Parmar, Uszkoreit, Jones,
  Gomez, Kaiser, and Polosukhin}]{vaswani2017attention}
Ashish Vaswani, Noam Shazeer, Niki Parmar, Jakob Uszkoreit, Llion Jones,
  Aidan~N Gomez, {\L}ukasz Kaiser, and Illia Polosukhin. 2017.
\newblock Attention is all you need.
\newblock In \emph{Proceedings of the 31st International Conference on Neural
  Information Processing Systems}, pages 6000--6010.

\bibitem[{Vyas et~al.(2014)Vyas, Gella, Sharma, Bali, and
  Choudhury}]{vyas2014pos}
Yogarshi Vyas, Spandana Gella, Jatin Sharma, Kalika Bali, and Monojit
  Choudhury. 2014.
\newblock Pos tagging of english-hindi code-mixed social media content.
\newblock In \emph{Proceedings of the 2014 Conference on Empirical Methods in
  Natural Language Processing (EMNLP)}, pages 974--979.

\bibitem[{Wang(2017{\natexlab{a}})}]{wang-2017-liar}
William~Yang Wang. 2017{\natexlab{a}}.
\newblock \href {https://doi.org/10.18653/v1/P17-2067} {{``}liar, liar pants on
  fire{''}: A new benchmark dataset for fake news detection}.
\newblock In \emph{Proceedings of the 55th Annual Meeting of the Association
  for Computational Linguistics (Volume 2: Short Papers)}, pages 422--426,
  Vancouver, Canada. Association for Computational Linguistics.

\bibitem[{Wang(2017{\natexlab{b}})}]{wang2017liar}
William~Yang Wang. 2017{\natexlab{b}}.
\newblock “liar, liar pants on fire”: A new benchmark dataset for fake news
  detection.
\newblock In \emph{Proceedings of the 55th Annual Meeting of the Association
  for Computational Linguistics (Volume 2: Short Papers)}, pages 422--426.

\bibitem[{Wharton Business~Daily(2020)}]{wharton}
Podcasts Wharton Business~Daily. 2020.
\newblock \href
  {https://knowledge.wharton.upenn.edu/article/how-social-media-is-shaping-political-campaigns/}
  {How social media is shaping political campaigns}.

\bibitem[{Wieling et~al.(2018)Wieling, Rawee, and van
  Noord}]{wieling2018reproducibility}
Martijn Wieling, Josine Rawee, and Gertjan van Noord. 2018.
\newblock Reproducibility in computational linguistics: Are we willing to
  share?
\newblock \emph{Computational Linguistics}, 44(4):641--649.

\bibitem[{Xue et~al.(2020)Xue, Constant, Roberts, Kale, Al-Rfou, Siddhant,
  Barua, and Raffel}]{xue2020mt5}
Linting Xue, Noah Constant, Adam Roberts, Mihir Kale, Rami Al-Rfou, Aditya
  Siddhant, Aditya Barua, and Colin Raffel. 2020.
\newblock mt5: A massively multilingual pre-trained text-to-text transformer.
\newblock \emph{arXiv preprint arXiv:2010.11934}.

\bibitem[{Yang et~al.(2019)Yang, Pentyala, Mohseni, Du, Yuan, Linder, Ragan,
  Ji, and Hu}]{yang2019xfake}
Fan Yang, Shiva~K Pentyala, Sina Mohseni, Mengnan Du, Hao Yuan, Rhema Linder,
  Eric~D Ragan, Shuiwang Ji, and Xia Hu. 2019.
\newblock Xfake: explainable fake news detector with visualizations.
\newblock In \emph{The World Wide Web Conference}, pages 3600--3604.

\bibitem[{Zhao et~al.(2021)Zhao, Da, and Yan}]{ZHAO2021102390}
Yuehua Zhao, Jingwei Da, and Jiaqi Yan. 2021.
\newblock \href {https://doi.org/https://doi.org/10.1016/j.ipm.2020.102390}
  {Detecting health misinformation in online health communities: Incorporating
  behavioral features into machine learning based approaches}.
\newblock \emph{Information Processing \& Management}, 58(1):102390.

\end{thebibliography}
\flushend

\end{document}